\documentclass{article}

\usepackage[final]{neurips_2019}

\usepackage[utf8]{inputenc} 
\usepackage[T1]{fontenc}    
\usepackage{hyperref}       
\usepackage{url}            
\usepackage{booktabs}       
\usepackage{amsfonts}       
\usepackage{amsmath}
\usepackage{amssymb}
\usepackage{bbm}
\usepackage{nicefrac}       
\usepackage{microtype}      
\usepackage{times}
\usepackage{graphicx} 
\usepackage[ruled,vlined]{algorithm2e}
\usepackage{multirow}

\newcommand{\method}{Sibling Rivalry}
\newcommand{\mthd}{SR}

\newcommand{\state}{s}
\newcommand{\action}{a}
\newcommand{\nextstate}{s'}
\newcommand{\goal}{g}
\newcommand{\optimum}{o}
\newcommand{\antigoal}{\bar{\goal}}
\newcommand{\State}{S}
\newcommand{\Action}{A}
\newcommand{\Goal}{G}

\newcommand{\reward}{r}
\newcommand{\Reward}{r}

\newcommand{\hp}{\epsilon}

\newcommand{\sreward}{\tilde{\reward}}
\newcommand{\sReward}{\tilde{\Reward}}

\newcommand{\fReward}{\Reward'}

\newcommand{\policy}{\pi}

\newcommand{\traj}{\boldsymbol{\tau}}
\newcommand{\that}{\traj^{f}}
\newcommand{\tbar}{\traj^{c}}
\newcommand{\rhohat}{\rho^{f}}
\newcommand{\rhobar}{\rho^{c}}

\newcommand{\sgmap}[1]{m({#1})}

\newcommand{\distance}[2]{d({#1}, {#2})}
\newcommand{\dthresh}{\delta}

\newcommand{\E}{\mathbb{E}}

\newcommand{\Rscalar}{\mathbb{R}}

\usepackage[dvipsnames]{xcolor}

\title{
Keeping Your Distance:
Solving Sparse Reward Tasks Using Self-Balancing Shaped Rewards
}

\author{%
  Alexander Trott \\
  Salesforce Research\\
  \texttt{atrott@salesforce.com} \\
  \And
  Stephan Zheng \\
  Salesforce Research\\
  \texttt{stephan.zheng@salesforce.com} \\
  \And
  Caiming Xiong \\
  Salesforce Research\\
  \texttt{cxiong@salesforce.com} \\
  \And
  Richard Socher \\
  Salesforce Research\\
  \texttt{rsocher@salesforce.com} \\
}

\begin{document}

\maketitle

\begin{abstract}
While using shaped rewards can be beneficial when solving sparse reward tasks, their successful application often requires careful engineering and is problem specific. 
For instance, in tasks where the agent must achieve some goal state, simple distance-to-goal reward shaping often fails, as it renders learning vulnerable to local optima.
We introduce a simple and effective model-free method to learn from shaped distance-to-goal rewards on tasks where success depends on reaching a goal state.
Our method introduces an auxiliary distance-based reward based on \textit{pairs} of rollouts to encourage diverse exploration.
This approach effectively prevents learning dynamics from stabilizing around local optima induced by the naive distance-to-goal reward shaping and enables policies to efficiently solve sparse reward tasks.
Our augmented objective does not require any additional reward engineering or domain expertise to implement and converges to the original sparse objective as the agent learns to solve the task.
We demonstrate that our method successfully solves a variety of hard-exploration tasks
(including maze navigation and 3D construction in a Minecraft environment),
where naive distance-based reward shaping otherwise fails, and intrinsic curiosity and reward relabeling strategies exhibit poor performance.
\end{abstract}


\section{Introduction}
\label{sec:introduction}


%
Reinforcement Learning (RL) offers a powerful framework for teaching an agent to perform tasks using only observations from its environment.
Formally, the goal of RL is to learn a policy that will maximize the reward received by the agent; for many real-world problems, this requires access to or engineering a reward function that aligns with the task at hand.
Designing a well-suited \textit{sparse} reward function simply requires defining the criteria for solving the task: reward is provided if the criteria for completion are met and withheld otherwise.
While designing a suitable sparse reward may be straightforward, learning from it within a practical amount of time often is not, often requiring exploration heuristics to help an agent discover the sparse reward \citep{Pathak2017, Burda2018, Burda2018a}.
\textit{Reward shaping} \citep{Mataric1994, Ng1999} is a technique to modify the reward signal, and, for instance, can be used to relabel and learn from failed rollouts, based on which ones made more progress towards task completion.
This may simplify some aspects of learning, but whether the learned behavior improves task performance depends critically on careful design of the shaped reward \citep{OpenAI_blog}.
As such, reward shaping requires domain-expertise and is often problem-specific \citep{Mataric1994}.

Tasks with well-defined goals provide an interesting extension of the traditional RL framework \citep{Kaelbling1993, Sutton2011, Schaul2015}.
Such tasks often require RL agents to deal with goals that vary across episodes and define success as achieving a state within some distance of the episode's goal.
Such a setting naturally defines a sparse reward that the agent receives when it achieves the goal.
Intuitively, the same distance-to-goal measurement can be further used for reward shaping (without requiring additional domain-expertise), given that it measures progress towards success during an episode.
%
%
However, reward shaping often introduces new local optima that can prevent agents from learning the optimal behavior for the original task. 
In particular, the existence and distribution of local optima strongly depends on the environment and task definition. 

As such, successfully implementing reward shaping quickly becomes problem specific.
%
%
These limitations have motivated the recent development of methods to enable learning from sparse rewards \citep{Schulman2017, Liu2019}, 
methods to learn latent representations that facilitate shaped reward \citep{Ghosh2018, Nair2018, Warde-Farley2019}, 
and learning objectives that encourage diverse behaviors \citep{Haarnoja2017, Eysenbach2018}.
%
%

We propose a simple and effective method to address the limitations of using distance-to-goal as a shaped reward.
%
%
%
In particular, we extend the naive distance-based shaped reward to handle \textit{sibling} trajectories, pairs of independently sampled trajectories using the same policy, starting state, and goal.
%
%
Our approach, which is simple to implement, can be interpreted as a type of self-balancing reward: we encourage behaviors that make progress towards the goal and simultaneously use sibling rollouts to estimate the local optima and encourage behaviors that avoid these regions, effectively balancing exploration and exploitation.
This objective helps to \textit{de}-stabilize local optima without introducing new stable optima, preserving the task definition given by the sparse reward.
This additional objective also relates to the entropy of the distribution of terminal states induced by the policy; however, unlike other methods to encourage exploration \citep{Haarnoja2017}, our method is ``self-scheduling'' such that our proposed shaped reward converges to the sparse reward as the policy learns to reach the goal.

Our method combines the learnability of shaped rewards with the generality of sparse rewards, which we demonstrate through its successful application on a variety of environments that support goal-oriented tasks.
In summary, our contributions are as follows: 

\begin{itemize}
    \item We propose \method, a method for model-free, dynamic reward shaping that preserves optimal policies on sparse-reward tasks.
    \item We empirically show that \method ~enables RL agents to solve hard-exploration sparse-reward tasks, where baselines often struggle to learn. We validate in four settings, including continuous navigation and discrete bit flipping tasks as well as hierarchical control for 3D navigation and 3D construction in a demanding Minecraft environment.
\end{itemize}

\section{Preliminaries}
\label{sec:preliminaries}

Consider an agent that must learn to maximize some task reward through its interactions with its environment.
At each time point $t$ throughout an episode, the agent observes its state $\state_t \in \State$ and selects an action $\action_t \in \Action$ based on its policy $\policy(\action_t | \state_t)$, yielding a new state $\nextstate_t$ sampled according to the environment's transition dynamics $p(\nextstate_t | \state_t, \action_t)$ and an associated reward $\reward_t$ governed by the task-specific reward function $\reward(\state_t, \action_t, \nextstate_t)$.
Let $\traj = \{(\state_t, \action_t, \nextstate_t, \reward_t)\}^{T-1}_{t=0}$ denote the trajectory of states, actions, next states, and rewards collected during an episode of length $T$, where $T$ is determined by either the maximum episode length or some task-specific termination conditions.
The objective of the agent is to learn a policy that maximizes its expected cumulative reward:
$\E_{\traj \sim \policy, p}\left[\Sigma_t \gamma^t \reward_t\right]$.



\paragraph{Reinforcement Learning for Goal-oriented tasks.}
The basic RL framework can be extended to a more general setting where the underlying association between states, actions, and reward can change depending on the parameters of a given episode \citep{Sutton2011}.
From this perspective, the agent must learn to optimize a \textit{set} of potential rewards, exploiting the shared structure of the individual tasks they each represent.
This is applicable to the case of learning a \textit{goal-conditioned} policy $\policy(\action_t | \state_t, \goal)$.
Such a policy must embed a sufficiently generic understanding of its environment to choose whatever actions lead to a state consistent with the goal $\goal$ \citep{Schaul2015}.
This setting naturally occurs whenever a task is defined by some set of goals $\Goal$ that an agent must learn to reach when instructed.
Typically, each episode is structured around a specific goal $\goal \in \Goal$ sampled from the task distribution. 
%
%
%
In this work, we make the following assumptions in our definition of ``goal-oriented task'':
\begin{enumerate}
    \item The task defines a distribution over starting states and goals $\rho(\state_0, \goal)$ that are sampled to start each episode.
    \item Goals can be expressed in terms of states such that there exists a function $\sgmap{\state} : \State \rightarrow \Goal$ that maps state $\state$ to its equivalent goal.
    \item \textbf{$S\times G \rightarrow \mathbb{R}^+$} An episode is considered a success once the state is within some radius of the goal, 
    such that $\distance{\state}{\goal} \leq \dthresh$,
    where $\distance{x}{y}: \Goal \times \Goal \rightarrow \mathbb{R}^+$ is a distance function\footnote{A straightforward metric, such as $L_1$ or $L_2$ distance, is often sufficient to express goal completion.} and $\dthresh \in \mathbb{R}^+$ is the distance threshold.
    (Note: this definition is meant to imply that the distance function internally applies the mapping $m$ to any states that are used as input; we omit this from the notation for brevity.)
\end{enumerate}

This generic task definition allows for an equally generic sparse reward function $\Reward (\state, \goal)$:
\begin{align}
\Reward(\state, \goal) &=
\begin{cases}
    1, &\distance{\state}{\goal} \leq \dthresh\\
    0, &\text{otherwise}
\end{cases}
\end{align}
From this, we define $\reward_t \triangleq \Reward(\nextstate_t, \goal)$ so that reward at time $t$ depends on the state reached after taking action $a_t$ from state $\state_t$.
Let us assume for simplicity that an episode terminates when either the goal is reached or a maximum number of actions are taken.
This allows us to define a single reward for an entire trajectory considering only the terminal state, giving: $\reward_{\traj} \triangleq \Reward(\state_T, \goal)$, where $\state_T$ is the state of the environment when the episode terminates.
The learning objective now becomes finding a goal-conditioned policy that maximizes
$\E_{\traj \sim \policy, p, ~ \state_0,\goal\sim\rho}\left[\reward_{\traj}\right]$.

\section{Approach}
\label{sec:approach}

\paragraph{Distance-based shaped rewards and local optima.}
We begin with the observation that the distance function $d$ (used to define goal completion and compute sparse reward) may be exposed as a shaped reward without any additional domain knowledge:
\begin{align}
    \sReward(\state, \goal) = 
    \begin{cases}
        1, &\distance{\state}{\goal} \leq \dthresh\\
        -\distance{\state}{\goal}, &\text{otherwise}
    \end{cases},~~~~~~
    \sreward_{\traj} \triangleq \sReward(\state_T, \goal).
\end{align}

\begin{figure}[t!]
  \begin{center}
	\includegraphics[width=1.0\textwidth]{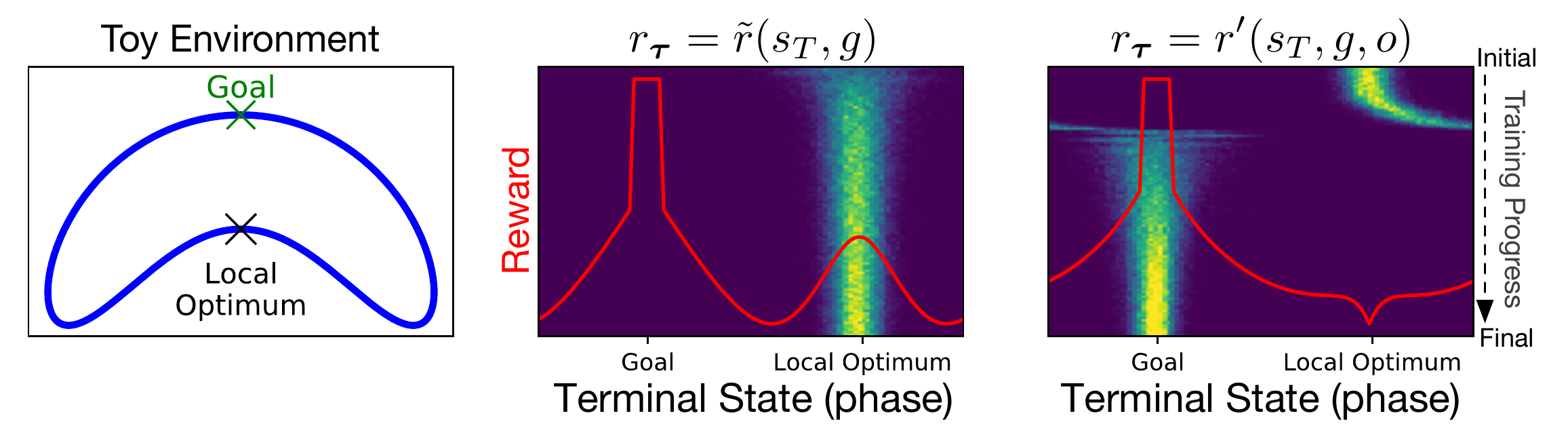}
  \end{center}
  \caption{
  \textbf{Motivating example.}
  (Left) The agent's task is to reach the goal (green \texttt{X}) by controlling its position along a warped circular track.
  A distance-to-goal reward ($L_2$ distance) creates a local optimum $\optimum_\goal$ (black \texttt{X}).
  (Middle and Right) Terminal state distributions during learning.
  The middle figure shows optimization using a distance-to-goal shaped reward.
  For the right figure, the shaped reward is augmented to include a hand-engineered bonus for avoiding $\optimum_\goal$ (Eq. \ref{eq:selfbalancingreward}; $\antigoal \leftarrow \optimum_\goal$).
  The red overlay illustrates the reward at each phase of the track.
  }
    \label{fig:figure_1_toy}
\end{figure}

By definition, a state that globally optimizes $\sReward$ also achieves the goal (and yields sparse reward), meaning that $\sReward$ preserves the global optimum of $\Reward$.
While we expect the distance function itself to have a single (global) optimum with respect to $\state$ and a fixed $\goal$, in practice we need to consider the possibility that other \textit{local} optima exist because of the state space structure, transition dynamics and other features of the environment.
For example, the agent may need to \emph{increase} its distance to the goal in order to eventually reach it.
This is exactly the condition faced in the toy task depicted in Figure \ref{fig:figure_1_toy}.
We would like to gain some intuition for how the learning dynamics are influenced by such local optima and how this influence can be mitigated.

The ``learning dynamics'' refer to the interaction between (i) the distribution of terminal states $\rho_{\goal}^{\policy}(\state_T)$ induced by a policy $\policy$ in pursuit of goal $\goal$ and (ii) the optimization of the policy with respect to $\E_{\rho_{\goal}^{\policy}(\state_T)}[ \sReward(\state_T, \goal)]$.
A local optimum $\optimum_\goal \in \State$
can be considered ``stable'' if, for all policies within some basin of attraction, continued optimization causes $\rho_{\goal}^{\policy}(\state_T)$ to converge to $\optimum_\goal$.
Figure \ref{fig:figure_1_toy} (middle) presents an example of this.
The agent observes its 2D position along the track and
takes an action to change its position;
its reward is based on its terminal state (after 5 steps).
Because of its starting position, maximizing the naive reward $\sReward{}(\state, \goal)$ causes the policy to ``get stuck'' at the local optimum $\optimum_\goal$,
i.e., the final state $\rho_{\goal}^{\policy}(\state_T)$ is peaked around $\optimum_\goal$.

In this example, the location of the local optimum is obvious and we can easily engineer a reward bonus for avoiding it.
In its more general form, this augmented reward is:
\begin{align}
    \fReward(\state, \goal, \antigoal) = 
    \begin{cases}
        1, &\distance{\state}{\goal} \leq \dthresh\\
        \min\left[ 0, -\distance{\state}{\goal} + \distance{\state}{\antigoal}\right], &\text{otherwise}
    \end{cases},~~~~~~
    \fReward_{\traj} \triangleq \fReward(\state_T, \goal, \antigoal).
    \label{eq:selfbalancingreward}
\end{align}

%
where $\antigoal \in \Goal$ acts as an `anti-goal' and specifies a state that the agent should avoid, e.g., the local optimum $\optimum_\goal$ in the case of the toy task in Figure \ref{fig:figure_1_toy}.
Indeed, using $\fReward$ and setting $\antigoal \leftarrow \optimum_\goal$ (that is, using $\optimum_\goal$ as the `anti-goal'), prevents the policy from getting stuck at the local optimum and enables the agent to quickly learn to reach the goal location (Figure \ref{fig:figure_1_toy}, right).

While this works well in this toy setting, the intuition for which state(s) should be used as the `anti-goal' $\antigoal$ will vary depending on the environment, the goal $g$, and learning algorithm.
In addition, using a fixed $\antigoal$ may be self-defeating if the resulting shaped reward introduces its own new local optima.
To make use of $\fReward(\state, \goal, \antigoal)$ in practice, we require a method to dynamically estimate the local optima that frustrate learning without relying on domain-expertise or hand-picked estimations.

\paragraph{Self-balancing reward.}
\begin{figure}[t!]
  \begin{center}
	\includegraphics[width=1.0\textwidth]{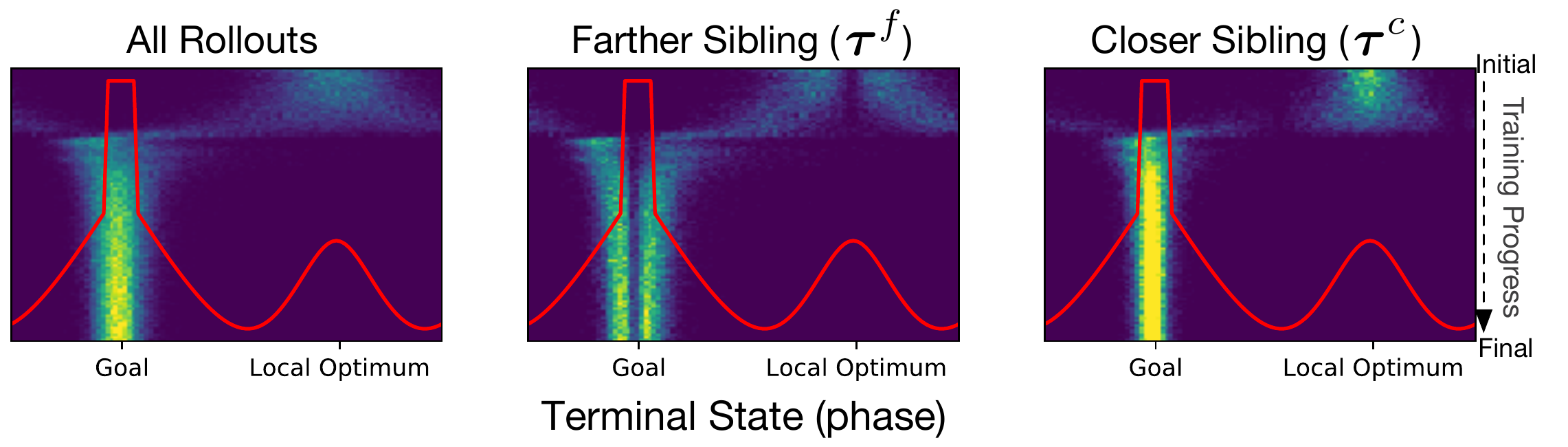}
  \end{center}
  \caption{
  \textbf{Learning with \method.}
  Terminal state distribution over training when using \mthd.
  Middle and right plots show the farther $\that$ and closer $\tbar$ trajectories, respectively.
  Red overlay illustrates the shape of the naive distance-to-goal reward $\sReward$.
  }
    \label{fig:figure_2_toy}
\end{figure}
We propose to estimate local optima directly from the behavior of the policy by using \textit{sibling} rollouts.
We define a pair of sibling rollouts as two independently sampled trajectories sharing the same starting state $\state_0$ and goal $\goal$.
We use the notation $\that, \tbar \sim \policy | \goal$ to denote a pair of trajectories from 2 sibling rollouts, where the superscript specifies that $\tbar$ ended closer to the goal than $\that$, i.e. that $\sreward_{\tbar} \geq \sreward_{\that}$.
By definition, optimization should tend to bring $\that$ closer towards $\tbar$ during learning.
That is, it should make $\that$ less likely and $\tbar$ more likely.
In other words, the terminal state of the closer rollout $\state_T^{c}$ can be used to estimate the location of local optima created by the distance-to-goal shaped reward.

To demonstrate this, we revisit the toy example presented in Figure \ref{fig:figure_1_toy} but introduce paired sampling to produce sibling rollouts (Figure \ref{fig:figure_2_toy}).
As before, we optimize the policy using $\fReward$ but with 2 important modifications.
First, we use the sibling rollouts for \textit{mutual relabeling} using the augmented shaped reward $\fReward$ (Eq. \ref{eq:selfbalancingreward}), where each rollout treats its sibling's terminal state as its own anti-goal:
\begin{align}
    \fReward_{\that} = \fReward(\state_T^{f}, \goal, \state_T^{c})
    \quad \& \quad
    \fReward_{\tbar} = \fReward(\state_T^{c}, \goal, \state_T^{f}).
\end{align}
Second, we only include the closer-to-goal trajectory $\tbar$ for computing policy updates if it reached the goal. 
As shown in the distribution of $\state_T^{c}$ over training (Figure \ref{fig:figure_2_toy}, right), $\state_T^{c}$ remains closely concentrated around \textit{an} optimum: the local optimum early in training and later the global optimum $\goal$.
Our use of sibling rollouts creates a reward signal that intrinsically balances exploitation and exploration by encouraging the policy to minimize distance-to-goal while de-stabilizing local optima created by that objective.
Importantly, as the policy converges towards the \textit{global} optimum (i.e. learns to reach the goal), $\fReward$ converges to the original underlying sparse reward $\Reward$.

\paragraph{\method.}

\begin{algorithm}[t!]
\label{algorithm}
\DontPrintSemicolon
\textbf{Given}
\begin{itemize}
    \item Environment, Goal-reaching task w/ $\State, \Goal, \Action, \rho(\state_0, \goal), \sgmap{}, \distance{}{}, \delta$ and max episode length\;
    \item Policy $\policy: \State \times \Goal \times \Action \rightarrow [0, 1]$ and Critic $V: \State \times \Goal \times \Goal \rightarrow \mathbb{R}$ with parameters $\theta$
    \item On-policy learning algorithm $\mathbb{A}$, e.g., REINFORCE, Actor-critic, PPO \;
    \item Inclusion threshold $\hp$
\end{itemize}
\For{iteration = 1...K}{
    Initialize transition buffer $D$\;
    \For{episode = 1...M}{
         Sample $\state_0, \goal \sim \rho$\;
         $\traj^a \leftarrow \policy_\theta(...) |_{\state_0, \goal}$ ~~~~~\# Collect rollout\;
         $\traj^b \leftarrow \policy_\theta(...) |_{\state_0, \goal}$ ~~~~~\# Collect sibling rollout\;
         Relabel $\traj^a$ reward using $\fReward$ and $\antigoal \leftarrow \sgmap{\state_T^b}$\;
         Relabel $\traj^b$ reward using $\fReward$ and $\antigoal \leftarrow \sgmap{\state_T^a}$\;
         \eIf{
            $\distance{\state_T^a}{\goal} < \distance{\state_T^b}{\goal}$
            }{
            $\tbar \leftarrow \traj^a$\;
            $\that \leftarrow \traj^b$\;
            }{
            $\tbar \leftarrow \traj^b$\;
            $\that \leftarrow \traj^a$\;
         }
         \eIf{
           $\distance{\state_T^c}{\state_T^f} < \hp$
           ~~\textbf{or}~~
           $\distance{\state_T^c}{\goal} < \delta$
           }{
           Add $\that$ and $\tbar$ to buffer $D$\;
           }{
           Add $\that$ to buffer $D$\;
         }
     }
     Apply on-policy algorithm $\mathbb{A}$ to update $\theta$ using examples in $D$\;
}
\caption{\method}
\end{algorithm}


From this, we derive a more general method for learning from sibling rollouts: \method{} (\mthd{}).
Algorithm \ref{algorithm} describes the procedure for integrating \mthd ~into existing on-policy algorithms for learning in the settings we described above.
\mthd{} has several key features:
\begin{enumerate}
    \item sampling sibling rollouts,
    \item mutual reward relabeling based on our self-balancing reward $\fReward$,
    \item selective exclusion of $\tbar$ (the closer rollout) trajectories from gradient estimation, using hyperparameter $\hp \in \Rscalar^+$ for controlling the inclusion/exclusion criterion.
\end{enumerate}
%

Consistent with the intuition presented above, we find that ignoring $\tbar$ during gradient estimation helps prevent the policy from converging to local optima.
In practice, however, it can be beneficial to learn directly from $\tbar$.
The hyperparameter $\hp$ serves as an inclusion threshold for controlling when $\tbar$ is included in gradient estimation, such that \mthd ~always uses $\that$ for gradient estimation and includes $\tbar$ only if it reaches the goal or if $\distance{\state_T^{f}}{\state_T^c} \leq \hp$.

The toy example above (Figure \ref{fig:figure_2_toy}) shows an instance of using \mthd ~where the base algorithm is A2C, the environment only yields end-of-episode reward ($\gamma=1$), and the closer rollout $\tbar$ is only used in gradient estimation when that rollout reaches the goal ($\hp=0$).
In our below experiments we mostly use end-of-episode rewards, although \mthd ~does not place any restriction on this choice.
It should be noted, however, that our method does require that full-episode rollouts are sampled in between parameter updates (based on the choice of treating the \textit{terminal} state of the sibling rollout as $\antigoal$) and that experimental control over episode conditions ($\state_0$ and $\goal$) is available.\footnote{Though we observe \mthd ~to work when $\state_0$ is allowed to differ between sibling rollouts (appendix, Sec. \ref{sec:appendix-start-state})} 
Lastly, we point out that we include the state $\state_t$, episode goal $\goal$, and anti-goal $\antigoal$ as inputs to the critic network $V$; the policy $\policy$ sees only $\state_t$ and $\goal$.

%
In the appendix, we present a more formal motivation of the technique (Section \ref{sec:appendix-motivation}), additional clarifying examples addressing the behavior of \mthd ~at different degrees of local optimum severity (Section \ref{sec:appendix-examples}), and an empirical demonstration (Section \ref{sec:appendix-hp-control}) showing how $\hp$ can be used to tune the system towards exploration ($\downarrow\hp$) or exploitation ($\uparrow\hp$).

\section{Experiments}
\label{sec:experiments}

To demonstrate the effectiveness of our method, we apply it to a variety of goal-reaching tasks.
We focus on settings where local optima interfere with learning from naive distance-to-goal shaped rewards.
We compare this baseline to results using our approach as well as to results using curiosity and reward-relabeling in order to learn from sparse rewards.
The appendix (Section \ref{sec:appendix-implementation}) provides detailed descriptions of the environments, tasks, and implementation choices.


\begin{figure}[t!]
  \begin{center}
	\includegraphics[width=1.0\textwidth]{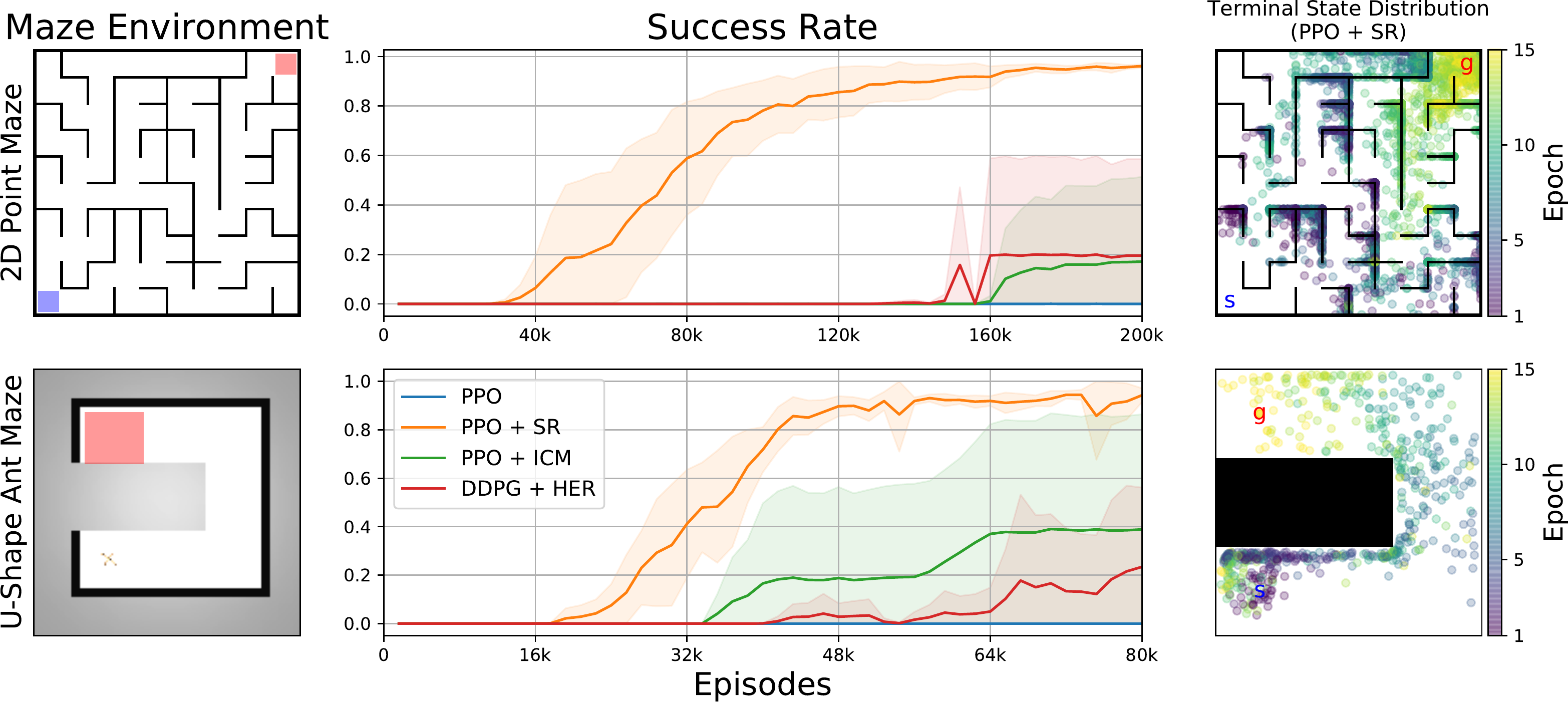}
  \end{center}
  \caption{
  \textbf{(Left) Maze environments.} Top row illustrates our 2D point maze; bottom row shows the U-shaped Ant Maze in a Mujoco environment.
  For the 2D maze, start location is sampled within the blue square; in the ant maze, the agent starts near its pictured location.
  For both, the goal is randomly sampled from within the red square region.
  \textbf{(Middle) Learning progress.} Lines show rate of goal completion averaged over 5 experiments (shaded area shows mean$\pm$SD, clipped to [0, 1]).
  Only our method (PPO+\mthd) allows the agent to discover the goal in all experiments. Conversely, PPO with the naive distance-to-goal reward never succeeds.
  Methods to learn from sparse rewards (PPO+ICM and DDPG+HER) only rarely discover the goals.
  Episodes have a maximum duration of 50 and 500 environment steps for the 2D Point Maze and Ant Maze, respectively.
  \textbf{(Right) State distribution.} Colored points illustrate terminal states achieved by the policy after each of the first 15 evaluation checkpoints. PPO+\mthd ~allows the agent to discover increasingly good optima without becoming stuck in them.
  }
    \label{fig:figure-pointmaze-antmaze}
\end{figure}


\paragraph{2D Point-Maze Navigation.}
How do different training methods handle the exploration challenge that arises in the presence of numerous local optima?
To answer this, we train an agent to navigate a fully-continuous 2D point-maze with the configuration illustrated in Figure \ref{fig:figure-pointmaze-antmaze} (top left).
At each point in time, the agent only receives its current coordinates and the goal coordinates.
It outputs an action that controls its change in location; the actual change is affected by collisions with walls.
When training using Proximal Policy Optimization \citep{Schulman2017} and a shaped distance-to-goal reward, the agent consistently learns to exploit the corridor at the top of the maze but never reaches the goal.
Through incorporating \method ~(PPO + \mthd), the agent avoids this optimum (and all others) and discovers the path to the goal location, solving the maze.

We also examine the behavior of algorithms designed to enable learning from sparse rewards without reward shaping.
Hindsight Experience Replay (HER) applies off-policy learning to relabel trajectories based on achieved goals \citep{Andrychowicz2017}.
In this setting, HER [using a DDPG backbone \citep{Lillicrap2016}] only learns to reach the goal on 1 of the 5 experimental runs, suggesting a failure in exploration since the achieved goals do not generalize to the task goals.
Curiosity-based intrinsic reward (ICM), which is shown to maintain a curriculum of exploration \citep{Pathak2017, Burda2018a}, fails to discover the sparse reward at the same rate.
Using random network distillation \citep{Burda2018}, a related intrinsic motivation method, the agent never finds the goal (not shown for visual clarity).
Only the agent that learns with \mthd ~is able to consistently and efficiently solve the maze (Figure \ref{fig:figure-pointmaze-antmaze}, top middle).



\paragraph{Ant-Maze Navigation using Hierarchical RL.}
\mthd ~easily integrates with HRL, which can help to solve more difficult problems such as navigation in a complex control environment \citep{Nachum2018}.
We use HRL to solve a U-Maze task with a Mujoco \citep{Todorov2012} ant agent (Figure \ref{fig:figure-pointmaze-antmaze}, bottom left), requiring a higher-level policy to propose subgoals based on the current state and the goal of the episode as well as a low-level policy to control the ant agent towards the given subgoal.
For fair comparison, we employ a standardized approach for training the low-level controller from subgoals using PPO but vary the approach for training the high-level controller.
For this experiment, we restrict the start and goal locations to the opposite ends of the maze (Figure \ref{fig:figure-pointmaze-antmaze}, bottom left).

The results when learning to navigate the ant maze corroborate those in the toy environment:
learning from the naive distance-to-goal shaped reward $\sReward$ fails because the wall creates a local optimum that policy gradient is unable to escape (PPO).
As with the 2D Point Maze, \mthd ~can exploit the optimum without becoming stuck in it (PPO+\mthd).
This is clearly visible in the terminal state patterns over early training (Figure \ref{fig:figure-pointmaze-antmaze}, bottom right).
We again compare with methods to learn from sparse rewards, namely HER and ICM.
As before, ICM stochastically discovers a path to the goal but at a low rate (2 in 5 experiments).
In this setting, HER struggles to generalize from its achieved goals to the task goals, perhaps due in part to the difficulties of off-policy HRL \citep{Nachum2018}.
3 of the 5 HER runs eventually discover the goal but do not reach a high level of performance.


\begin{figure}[t!]
  \begin{center}
	\includegraphics[width=1.0\textwidth]{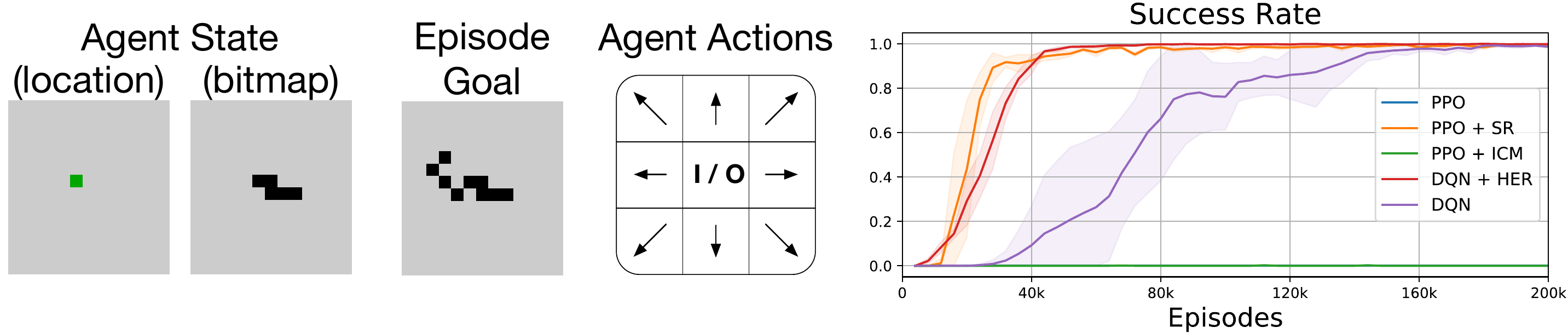}
  \end{center}
  \caption{
  \textbf{2D discrete pixel-grid environment.}
  The agent begins in a random location on a 13x13 grid with all pixels off and must move and toggle pixels to produce the goal bitmap.
  The agent sees its current location (1-hot), the current bitmap, and the goal bitmap.
  The agent succeeds when the bitmap exactly matches the goal (0-distance).
  Lines show rate of goal completion averaged over 5 experiments (shaded area shows mean$\pm$SD, clipped to [0, 1]).
  Episodes have a maximum duration of 50 environment steps.
  }
    \label{fig:figure-pixgrid}
\end{figure}

\paragraph{Application to a Discrete Environment.}
Distance-based rewards can create local optima in less obvious settings as well.
To examine such a setting and to show that our method can apply to environments with discrete action/state spaces, we experiment with learning to manipulate a 2D bitmap to produce a goal configuration.
The agent starts in a random location on a 13x13 grid and may move to an adjacent location or toggle the state of its current position (Figure \ref{fig:figure-pixgrid}, left).
We use $L_1$ distance (that is, the sum of bitwise absolute differences).
Interestingly, this task does not require the agent to increase the distance to the goal in order to reach it (as, for example, with the Ant Maze), but naive distance-to-goal reward shaping still creates `local optima` by introducing pathological learning dynamics:
early in training, when behavior is closer to random, toggling a bit from off to on tends to \textit{increase} distance-to-goal and the agent quickly learns to avoid taking the toggle action.
Indeed, the agents trained with naive distance-to-goal reward shaping $\sReward$ never make progress (PPO).
As shown in Figure \ref{fig:figure-pixgrid}, we can prevent this outcome and allow the agent to learn the task through incorporating \method ~(PPO+\mthd).

As one might expect, off-policy methods that can accommodate forced exploration may avoid this issue;
DQN \citep{Mnih2015} gradually learns the task (note: this required densifying the reward rather than using only the terminal state).
However, exploration alone is not sufficient on a task like this since simply achieving diverse states is unlikely to let the agent discover the task structure relating states, goals, and rewards, as evidenced by the failure of ICM to enable learning in this setting.
HER aims to learn this task structure from failed rollouts and, as an off-policy method, handles forced exploration, allowing it to quickly learn this task.
Intuitively, using distance as a reward signal automatically exposes the task structure but often at the cost of unwanted local optima.
\method ~avoids that tradeoff, allowing efficient on-policy learning\footnote{We find that including both sibling trajectories ($\hp=\texttt{Inf}$) works best in the discrete-distance settings}.

\begin{figure}[t!]
  \begin{center}
	\includegraphics[width=1.0\textwidth]{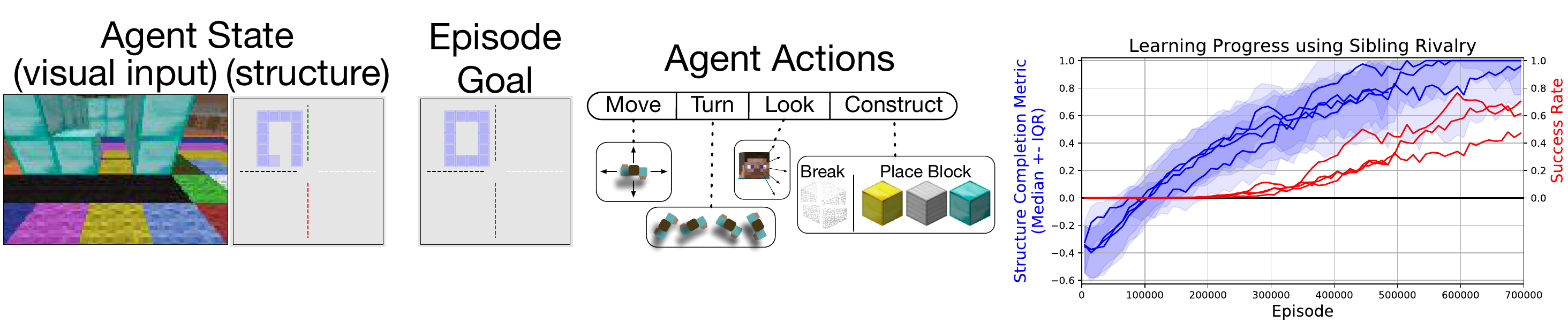}
  \end{center}
  \caption{
  \textbf{3D construction task in Minecraft.}
  The agent must control its location/orientation and break/place blocks in order to produce the goal structure.
  The agent observes its first-person visual input, the discrete 3D cuboid of the construction arena, and the corresponding cuboid of the goal structure.
  An episode is counted as a success when the structure exactly matches the goal.
  The \textit{Structure Completion Metric} is difference between correctly and incorrectly placed blocks divided by the number of goal-structure blocks.
  In the illustrated example, the agent has nearly constructed the goal, which specifies a height-2 diamond structure near the top left of the construction arena.
  Goal structures vary in height, dimensions, and material (4806 unique combinations).
  Episodes have a maximum duration of 100 environment steps.
  }
    \label{fig:minecraft}
\end{figure}

\paragraph{3D Construction in Minecraft.}
Finally, to demonstrate that \method ~can be applied to learning in complex environments, we apply it to a custom 3D construction task in Minecraft using the Malmo platform \citep{Johnson2016}.
Owing to practical limitations, we use this setting to illustrate the scalability of \mthd ~rather than to provide a detailed comparison with other methods.
Similar to the pixel-grid task, here the agent must produce a discrete goal structure by placing and removing blocks (Figure \ref{fig:minecraft}).
However, this task introduces the challenge of a first-person 3D environment, combining continuous and discrete inputs, and application of aggressively asynchronous training with distributed environments [making use of the IMPALA framework \citep{Espeholt2018}].
Since success requires exact-match between the goal and constructed cuboids, we use the number of block-wise differences as our distance metric.
Using this distance metric as a naive shaped reward causes the agent to avoid ever placing blocks within roughly 1000 episodes (not shown for visual clarity).
Simply by incorporating \method ~the agent avoids this local optimum and learns to achieve a high degree of construction accuracy and rate of exact-match success (Figure \ref{fig:minecraft}, right).

\section{Related Work}
\label{sec:relatedwork}


\paragraph{Intrinsic Motivation.}
Generally speaking, the difficulty in learning from sparse rewards comes from the fact that they tend to provide prohibitively rare signal to a randomly initialized agent.
Intrinsic motivation describes a form of task-agnostic reward shaping that encourages exploration by rewarding novel states.
Count-based methods track how often each state is visited to reward reaching relatively unseen states \citep{Bellemare2016, Tang2017}.
Curiosity-driven methods encourage actions that surprise a separate model of the network dynamics \citep{Pathak2017, Burda2018a, Zhao2019}.
\citet{Burda2018} introduce a similar technique using distillation of a random network.
In addition to being more likely to discover sparse reward, policies that produce diverse coverage of states provide a strong initialization for downstream tasks \citep{Haarnoja2017, Eysenbach2018}.
Intrinsic motivation requires that the statistics of the agent's experience be directly tracked or captured in the training progress of some external module.
In contrast, we use the policy itself to estimate and encourage exploratory behavior.

\paragraph{Curriculum Learning and Self-Play.}
Concepts from curriculum learning \citep{Bengio2009} have been applied to facilitate learning goal-directed tasks \citep{Molchanov2018, Nair2018}.
\citet{Florensa2018}, for example, introduce a generative adversarial network approach for automatic generation of a goal curriculum.
On competitive tasks, such as 2-player games, self-play has enabled remarkable success \citep{Silver2018}.
Game dynamics yield balanced reward and force agents to avoid over-committing to suboptimal strategies, providing both a natural curriculum and incentive for exploration.
Similar benefits have been gained through asymmetric self-play with goal-directed tasks \citep{Sukhbaatar2018a, Sukhbaatar2018}.
Our approach shares some inspiration with this line of work but combines the asymmetric objectives into a single reward function.

\paragraph{Learning via Generalization.}
Hindsight Experience Replay \citep{Andrychowicz2017} combines reward relabeling and off-policy methods to allow learning from sparse reward even on failed rollouts, leveraging the generalization ability of neural networks as universal value approximators \citep{Schaul2015}.
Asymmetric competition has been used to improve this method, presumably by inducing an automatic exploration curriculum that helps relieve the generalization burden \citep{Liu2019}.

\paragraph{Latent Reward Shaping.}
A separate approach within reward shaping involves using latent representations of goals and states.
\citet{Ghosh2018} estimate distance between two states based on the actions a pre-trained policy would take to reach them.
\citet{Nair2018} introduce a method for unsupervised learning of goal spaces that allows practicing reaching imagined goal states by computing distance in latent space [see also \citet{Pere2018}].
\citet{Warde-Farley2019} use discriminitive training to learn to estimate similarity to a goal state from raw observations.
\section{Conclusion}
\label{sec:conclusion}

We introduce \method, a simple and effective method for learning goal-reaching tasks from a generic class of distance-based shaped rewards.
\method ~makes use of sibling rollouts and self-balancing rewards to prevent the learning dynamics from stabilizing around local optima.
By leveraging the distance metric used to define the underlying sparse reward, our technique enables robust learning from shaped rewards without relying on carefully-designed, problem-specific reward functions.
We demonstrate the applicability of our method across a variety of goal-reaching tasks where naive distance-to-goal reward shaping consistently fails and techniques to learn from sparse rewards struggle to explore properly and/or generalize from failed rollouts. 
Our experiments show that \method ~can be readily applied to both continuous and discrete domains, incorporated into hierarchical RL, and scaled to demanding environments.


\bibliography{bib}
\bibliographystyle{bib_style}

\appendix

\section{Formal Motivation}
\label{sec:appendix-motivation}

Here, we present a hypothesis relating \textit{sibling rollouts} (that is, independently sampled rollouts using the same policy $\policy$, starting state $\state_0$, and goal $\goal$) to the learning dynamics created by distance-to-goal shaped rewards $\sReward$.
%
%
%
%
We use the notation $\that, \tbar \sim \policy | \goal$ to denote a pair of trajectories from 2 sibling rollouts, where the superscript specifies that $\tbar$ ended closer to the goal than $\that$, i.e. that $\sreward_{\tbar} \geq \sreward_{\that}$.
We use $\phi_{\goal}^{\policy}(\traj)$ to denote the probability that trajectory $\traj$ earns a higher reward (i.e., is closer to the goal) than a sibling trajectory $\traj'$
\begin{align}
    \phi_{\goal}^{\policy}(\traj) = \E_{\traj' \sim \policy|\goal}\left[ 1(\sreward_{\traj} > \sreward_{\traj'}) \right].
\end{align}
This allows us to define the marginal (un-normalized) distributions for the sibling trajectories $\tbar$ and $\that$ as
\begin{align}
    \rhobar(\traj | \goal) &= \policy(\traj | \goal) \cdot \phi_{\goal}^{\policy}(\traj),\\
    \rhohat(\traj | \goal) &= \policy(\traj | \goal) \cdot (1 - \phi_{\goal}^{\policy}(\traj)),
\end{align}
where $\policy(\traj|\goal) = \sum_{\action, \state \in \traj}\policy(\action | \state, \goal)$.

Let us define the \textit{pseudoreward} $\psi_{\goal}^{\policy}(\traj)$ as a simple scaling and translation of $\phi_{\goal}^{\policy}(\traj)$:
\begin{align}
    \psi_{\goal}^{\policy}(\traj) = 2\phi_{\goal}^{\policy}(\traj) - 1.
\end{align}
Importantly, since $\psi_{\goal}^{\policy}(\traj)$ captures how $\traj$ compares to the distribution of trajectories induced by $\policy$, the policy can always improve its expected reward by increasing the probability of $\traj$ for all $\traj$ where $\psi_{\goal}^{\policy}(\traj) > 0$ and decreasing the probability of $\traj$ for all $\traj$ where $\psi_{\goal}^{\policy}(\traj) < 0$.
Noting also that $\psi_{\goal}^{\policy}(\traj)$ increases monotonically with $\sReward_{\traj}$, we may gain some insight into the learning dynamics when optimizing $\sReward_{\traj}$ by considering the definition of the policy gradient for optimizing $\psi_{\goal}^{\policy}(\traj)$:
\begin{align}
    \nabla_\theta \E_{\traj \sim \policy|\goal}[\psi_{\goal}^{\policy}(\traj)]
    &= \E_{\traj \sim \policy|\goal}[\nabla_\theta \log \policy(\traj | \goal) \cdot (2\phi_{\goal}^{\policy}(\traj)-1)]\\
    &= \int \nabla_\theta \log \policy(\traj | \goal) \cdot \left[ \rhobar(\traj | \goal) - \rhohat(\traj | \goal) \right] d\traj,
\end{align}
where $\theta$ is the set of internal parameters governing $\policy$.

This comparison exposes the importance of the quantity $\rhobar(\traj | \goal) - \rhohat(\traj | \goal)$, suggesting that the gradients should serve to reinforce trajectories where this difference is maximized.
In other words, this suggests that optimizing $\policy$ with respect to $\sReward$ will concentrate the policy around trajectories that are \textit{over} represented in $\rhobar(\traj | \goal)$ and \textit{under} represented in $\rhohat(\traj | \goal)$.

While we cannot measure the marginal probabilities $\rhobar(\traj | \goal)$ and $\rhohat(\traj | \goal)$ for a given trajectory $\traj$, we can sample trajectories from these distributions via sibling rollouts.
\method ~applies the interpretation that samples from $\rhobar(\traj | \goal)$ (i.e., closer-to-goal siblings) capture the types of trajectories that the policy will converge towards when optimizing the distance-to-goal reward.
By both limiting the use of $\tbar$ trajectories when computing gradients and encouraging trajectories to avoid the terminal states achieved by their sibling, we can counteract the learning dynamics that cause the policy to converge to a local optimum.

%

\section{Additional Examples}
\label{sec:appendix-examples}

\begin{figure}[t!]
  \begin{center}
	\includegraphics[width=1.0\textwidth]{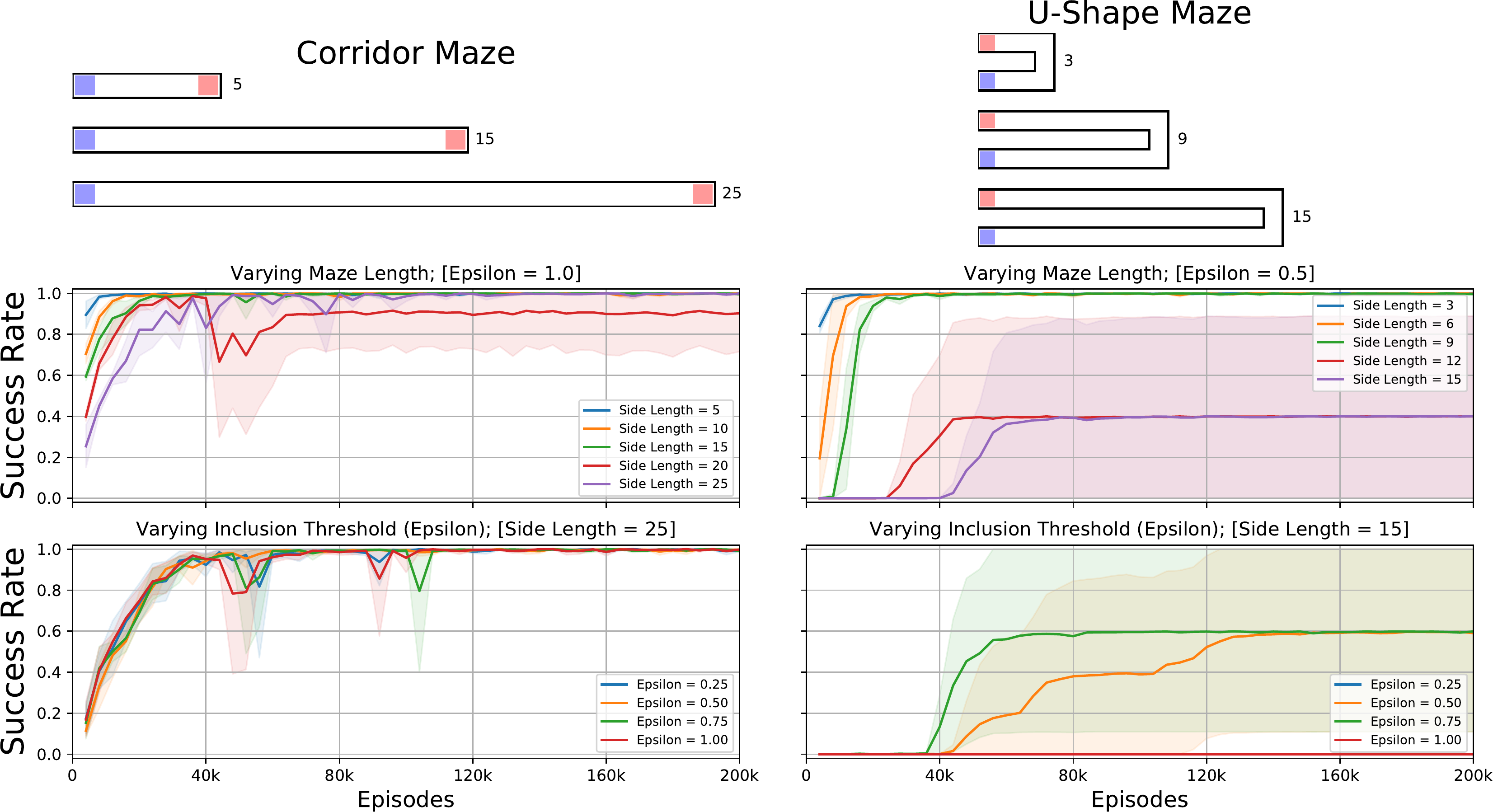}
  \end{center}
  \caption{
  \textbf{Performance and limitations of \method ~in toy 2D point mazes.}
  (Top) Corridor mazes (left) and U-shaped mazes (right) of varying side lengths.
  The number to the right of the maze indicates the side length.
  The start/goal location is sampled within the blue/red squares, respectively.
  (Middle) Performance of \mthd ~for each of the tested side lengths.
  (Bottom) Performance at varying settings of $\hp$, the inclusion hyperparameter, on the longest variant of the corridor (left) and U-shaped (right) mazes.
  Lines show rate of goal completion averaged over 5 experiments (shaded area shows mean$\pm$SD, clipped to [0, 1]).
  }
    \label{fig:pointmaze-corridor-umaze}
\end{figure}

To further illustrate the behavior of \method ~we profile learning in two simplified versions of the 2D point maze environment: an elongated corridor and a U-shaped hallway (Figure \ref{fig:pointmaze-corridor-umaze}).
In each, the agent starts at one end and must reach a goal location at the other end of the ``maze.''
These two variants allow us to examine any tension between the distance-to-goal and distance-from-antigoal (i.e. distance from sibling terminal state) components of the reward $\fReward$ used by \mthd.

In other words: \textbf{what is the trade-off between avoiding local optima created by the distance function and pursuing the global optimum created by the distance function?}
 
We address this question by comparing performance in corridor mazes and U-shaped mazes of varying side lengths (Figure \ref{fig:pointmaze-corridor-umaze}, Top \& Middle).
In the corridor maze, the distance-to-goal signal creates a single global optimum that the agent should pursue.
In the U-shaped maze, the distance-to-goal signal creates a local optimum (in addition to the global optimum) that the agent must avoid.
At the longest side-length tested, nearly all points within the U-shaped maze yield a worse distance-to-goal reward than the point at the local optimum, making the local optimum very difficult to escape.
It is worth noting here that curiosity (ICM) and HER were observed to fail on both of these maze variants for the longest tested side lengths.

As shown in Figure \ref{fig:pointmaze-corridor-umaze}, \mthd ~quickly solves the corridor maze, where the distance-to-goal signal serves as a good shaped reward.
This result holds at the longest corridor setting for each of the $\hp$ settings tested (Figure \ref{fig:pointmaze-corridor-umaze}, Bottom).
Note: these settings correspond to fairly aggressive exclusion of the closer-to-goal sibling rollout.
This simple environment offers a clear demonstration that \mthd ~preserves the distance-to-goal reward signal.
However, as discussed in Section \ref{sec:appendix-hp-control}, using an overly aggressive $\hp$ can lead to worse performance in a more complex environment.

Importantly, \mthd ~also solves the U-shaped variants, which are characterized by severe local optima.
However, while we still observe decent performance for the most difficult versions of the U-shaped maze, this success depends strongly on a carefully chosen setting for $\hp$.
\textit{As the distance function becomes increasingly misaligned with the structure of the environment, the range of good values for $\hp$ shrinks.}
Section \ref{sec:appendix-hp-control} provides further empirical insight into the influence of $\hp$.

The combined observations from the corridor and U-shaped mazes illustrate that \method ~achieves a targeted disruption of the learning dynamics associated with (non-global) local optima.
This explains why \mthd ~does not interfere with solving the corridor maze, where local optima are not an issue, while being able to solve the U-shaped maze, characterized by severe local optima.
Furthermore, these observations underscore that using $\fReward$ and sibling rollouts for reward re-labeling automatically tailors the reward signal to the environment/task being solved.

\section{Controlling the Inclusion Hyperparameter}
\label{sec:appendix-hp-control}

\begin{figure}[t!]
  \begin{center}
	\includegraphics[width=1.0\textwidth]{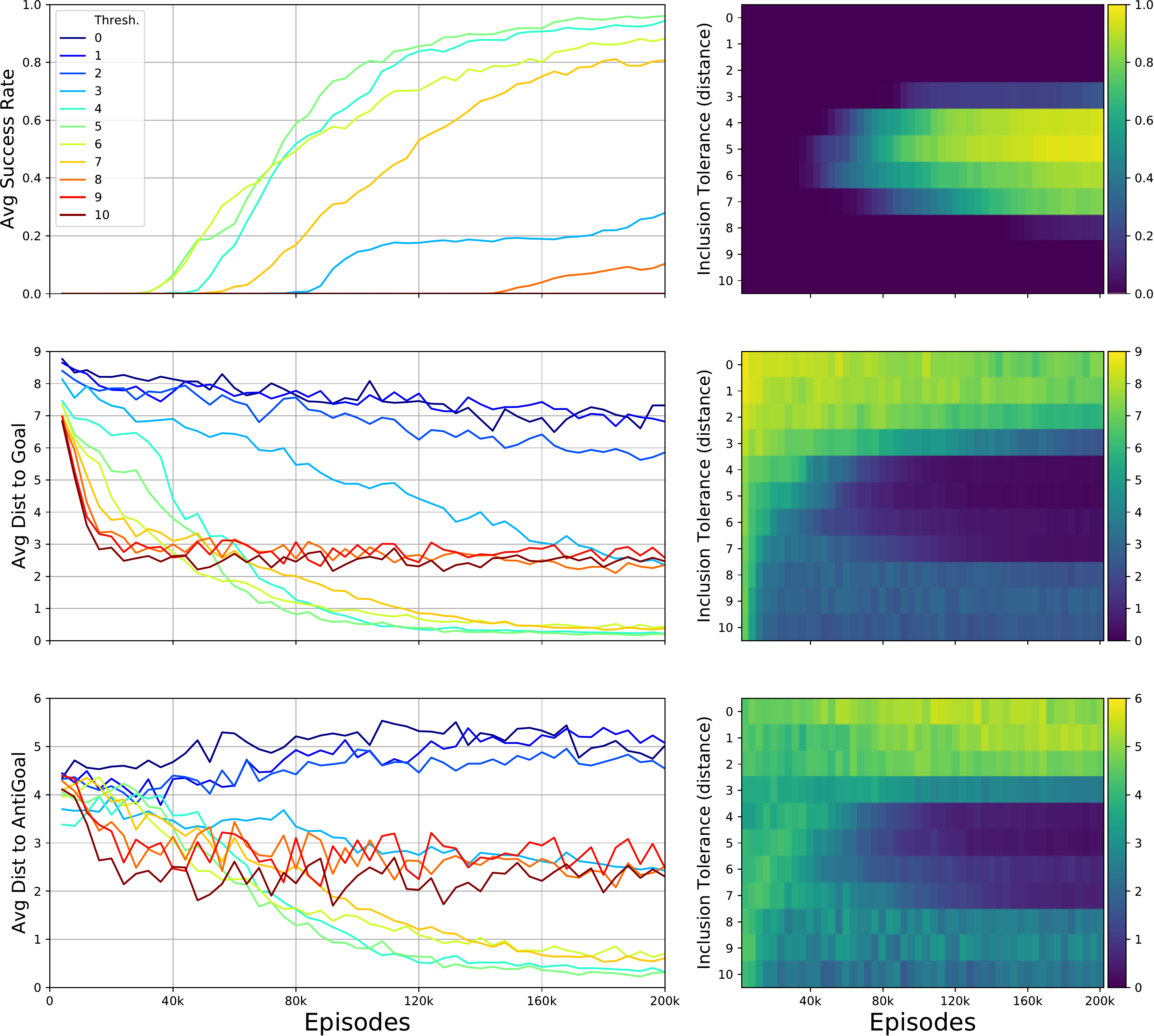}
  \end{center}
  \caption{
  \textbf{Effect of Inclusion Threshold $\hp$ on \method.} We re-run the 2D point maze experiments using \mthd ~with each of the $\hp$ settings shown.
  Rows report success rate, distance to goal, and distance to anti-goal (that is, distance between sibling rollouts) across training for each of the settings.
  Line plots and heatmap plots provide different views of the same data.
  This analysis identifies roughly 3 modes of behavior exhibited by our method in this environment.
  The first, over-exploration, occurs for the lower range of $\hp$, where closer-to-goal trajectories are more aggressively discarded.
  Close inspection shows slow progress towards the goal and a tendency to increase inter-sibling distance (the latter trend appears to reverse near the end of the training window).
  The second mode corresponds to successful behavior: the agent can exploit the distance-to-goal signal but maintains enough diversity in its state distribution to avoid commitment to local optima.
  The third mode, under-exploration, occurs for the higher range of $\hp$, where inclusion of the closer-to-goal trajectory is more permissive.
  These settings lead the agent to the same pitfall that prevents learning from naive distance-to-goal shaped rewards.
  That is, it quickly identifies a low-distance local optimum (consistently, the top corridor of the maze) and does not sufficiently explore in order to find a higher-reward region of the maze.
  }
    \label{fig:pointmaze-hp}
\end{figure}

\method ~makes use of a single hyperparameter $\hp$ to set the distance threshold for when to include the closer-to-goal trajectory $\tbar$ in the parameter updates.
When $\hp=0$, $\tbar$ is only included if it reaches the goal.
Conversely, when $\hp=\texttt{Inf}$, the algorithm always uses both trajectories (while still encouraging diversity through the augmented reward $\fReward$).
We find that this parameter can be used to tune learning towards exploration or exploitation (of the distance-to-goal reward).

This is most evident in the impact of $\hp$ on learning progress in the 2D point maze environment, where local optima are numerous (and, in our observation, learning progress is most sensitive to $\hp$).
For the sake of demonstration, we performed a set of experiments for each of $\hp \in [0, 1, ... 10]$ distance units.
The 2D point maze itself is 10x10, giving us good coverage of options one might consider for $\hp$ in this environment.
Interestingly, we observe three modes of the algorithm: over-exploration ($\hp$ too low), successful learning, and under-exploration ($\hp$ too high).
We observe these modes to be clearly identifiable using the metrics reported in Figure \ref{fig:pointmaze-hp}.
In practice a much coarser search over this hyperparameter should be sufficient to identify the optimal range.

\section{Robustness to Different Starting States}
\label{sec:appendix-start-state}

\begin{figure}[t!]
  \begin{center}
	\includegraphics[width=1.0\textwidth]{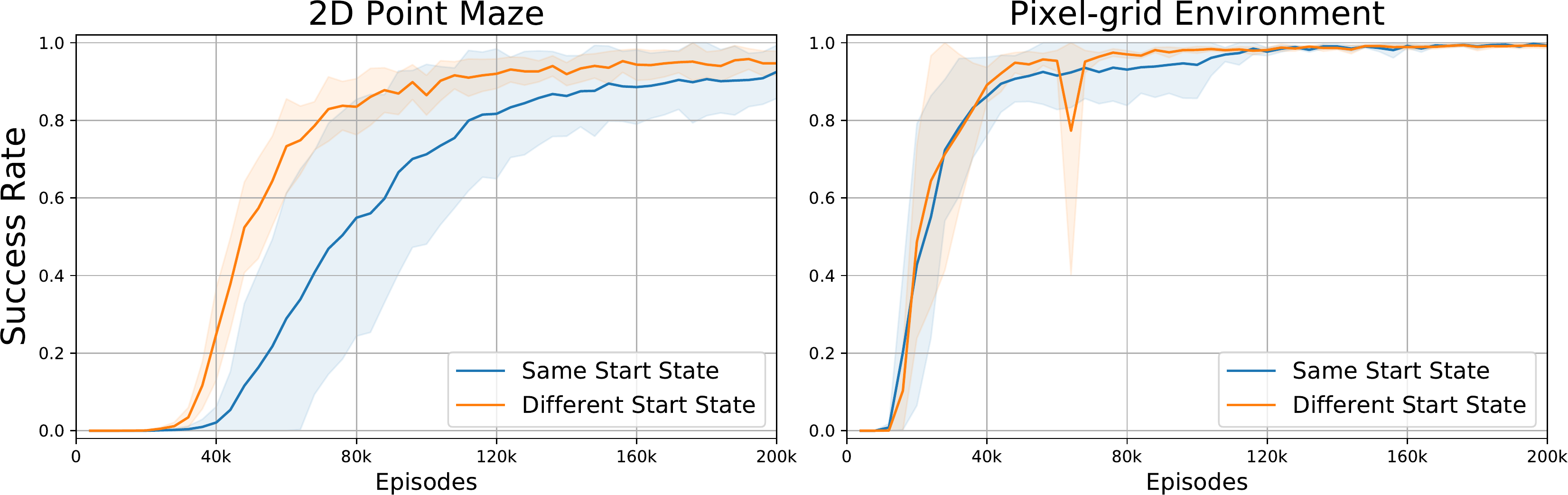}
  \end{center}
  \caption{
  \textbf{Robustness to Different Starting States.}
  Performance of \method ~when sibling rollouts share the same starting state $\state_0$ versus independently sampled starting states.
  Curves show training progress for the 2D point maze task (left) and for the pixel-grid bit flipping task (right), averaged over 5 experiments (shaded area shows mean$\pm$SD, clipped to [0, 1]).
  Blue curves (same start state) follow the definition of `sibling rollout` used in the main text.
  }
    \label{fig:start-state}
\end{figure}

Since some learning settings do not offer direct control over the starting state of an episode, we test the performance of \method ~when the start states of sibling rollouts are sampled independently (Figure \ref{fig:start-state}).
For the 2D point maze environment, start locations are sampled independently from within the bottom left corner of the maze.
For the pixel-grid environment, sibling rollouts use independently sampled grid locations as the starting position.
In both cases, the siblings' starting states correspond to 2 independent samples from the task's underlying start state distribution.
We compare performance under these sampling conditions to performance when the sibling rollouts use the same starting state.
Interestingly, we observe faster convergence with independent starting states for the 2D point maze and roughly similar performance for the pixel-grid environment.
These results suggest that \method ~is robust to noise in the starting state and may even benefit from it.
However, this is not an exhaustive analysis and one might expect different outcomes for environments where the policy tends to find different local optima depending on the episode's starting state.
Nevertheless, these results indicate that \method ~can be applied in settings where exact control over $\state_0$ is not feasible.

\section{Comparison to Grid Oracle Baseline}
\label{sec:appendix-grid-oracle}

\begin{figure}[t!]
  \begin{center}
	\includegraphics[width=0.8\textwidth]{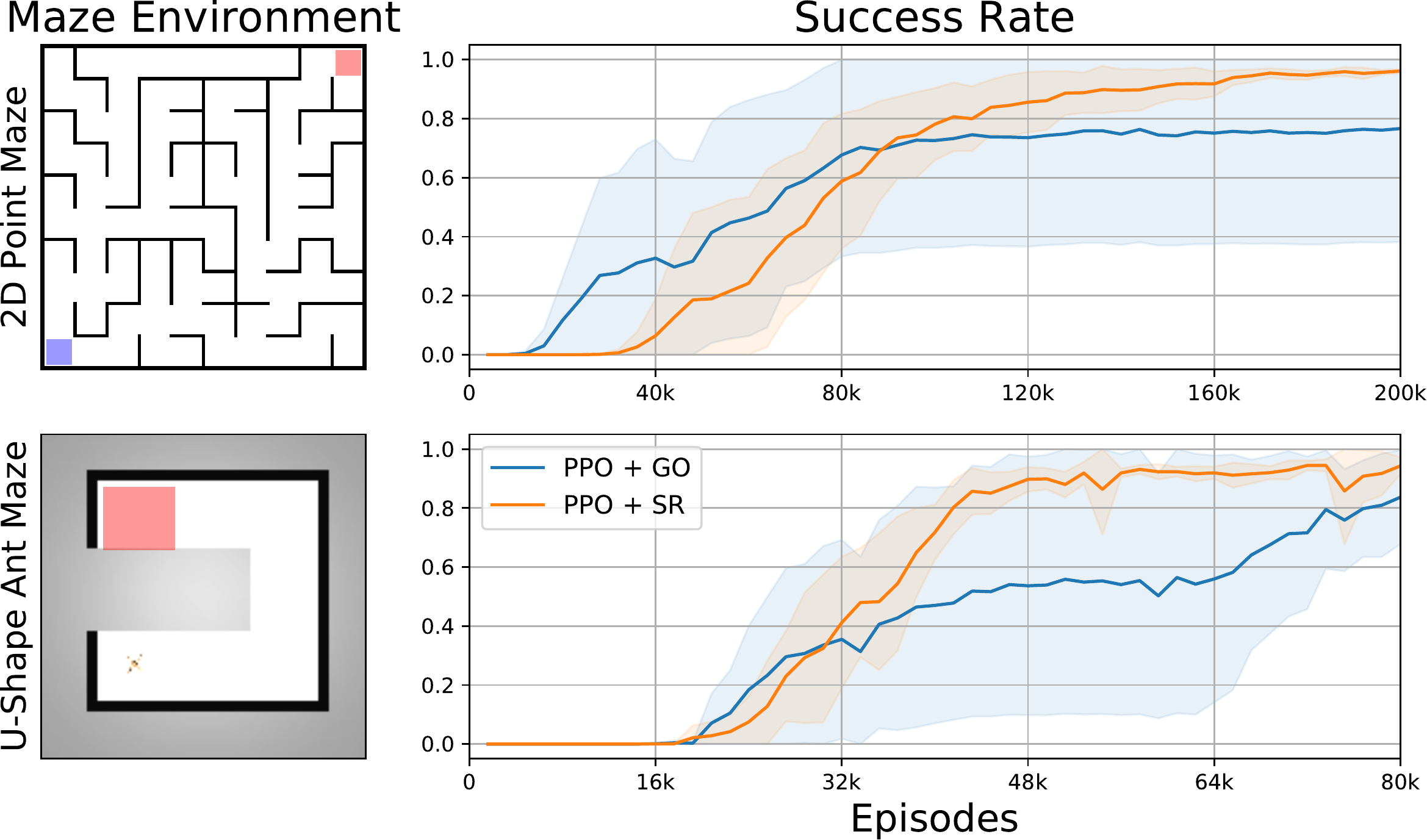}
  \end{center}
  \caption{
  \textbf{Comparison with Grid Oracle baseline.} 
  We compare performance of \mthd ~in the maze environments to performance when using a Grid Oracle reward.
  \mthd ~(PPO + SR) employs the self-balancing shaped reward $\fReward$ (described in the main text), whereas the Grid Oracle (PPO + GO) adds a shaped reward based on the number of discrete environment regions visited within an episode.
  For the 2D maze, the start location is sampled within the blue square; in the ant maze, the agent starts near its pictured location.
  For both, the goal is randomly sampled from within the red square region.
  Lines show rate of goal completion averaged over 5 experiments (shaded area shows mean$\pm$SD, clipped to [0, 1]).
  }
    \label{fig:gridoracle}
\end{figure}

We compare the performance of \method ~to a \textit{Grid Oracle} baseline \citep{Savinov2019}.
The Grid Oracle augments the end-of-episode reward with a value proportional to the number of regions visited during the episode, computed by dividing the XY-space of the environment into a grid of discrete regions (with the number of divisions serving as the main hyperparameter).
The Grid Oracle only sees the sparse reward plus the region-visitation reward.
This baseline encourages the agent to cover a broad area, which, based on the exploration challenge presented by the maze environments, may act as a generically useful shaped reward for helping to discover the sparse reward from reaching the goal.
Using the Grid Oracle shaped reward indeed facilitates discovery of the goal but can suffer from the fact that maximizing coverage within an episode does not guarantee task-useful behavior (Figure \ref{fig:gridoracle}).
Interestingly, we find that \mthd ~tends to more consistently solve the 2D point maze and ant maze tasks.
It is also worth noting that \mthd ~only slightly lags the Grid Oracle baseline in terms of sample complexity.
\mthd ~encourages both efficient and task-useful exploration by taking advantage of the properties of sibling rollouts and distance-based rewards.

\section{Implementation Details and Experimental Hyperparameters}
\label{sec:appendix-implementation}

Here, we provide a more detailed description of the environments, tasks, and training implementations used in our experiments (Section \ref{sec:experiments}).
We first provide a general description of the training algorithms as they pertain to our experiments.
We follow with task-specific details for each of the environments.

For all experiments, we distribute rollout collection over 20 parallel threads.
Quantities regarding rollouts, epochs, and minibatches are all reported \textit{per worker}.

\paragraph{Proximal Policy Optimization (PPO).}
Many of the experiments we perform use PPO as the backbone learning algorithm.
We focus on PPO because of its strong performance and because it is well suited for the constraints imposed by the application of \method.
Specifically, our method requires the collection of multiple full rollouts in between network updates.
PPO handles this well as it is able to make multiple updates from a large batch of transitions.
While experimental variants that do not use \mthd ~do not require scheduling updates according to full rollouts, we do so for ease of comparison.
The general approach we employ cycles between collection of full trajectories and multiple optimization epochs over minibatches of transitions within those trajectories.
We apply a constant number of optimization epochs and updates per epoch, varying the sizes of the minibatches as needed based on the variable length of trajectories (due to either episode termination after goal-reaching or trajectory exclusion when using \mthd).
We confirmed that this modification of the original algorithm did not meaningfully affect learning.

We standardize our PPO approach as much as possible to avoid results due to edge-case hyperparameter configurations, using manual search to identify such generally useful parameter settings.
In the ant maze task, this standardized approach applies specifically to training the high-level policy.
We also use PPO to train the low-level policy but adopt a more specific approach for that based on its unique role in our experiments (described below).

For PPO variants, the output head of the policy network specifies the $\alpha \in \Rscalar^2$ and $\beta \in \Rscalar^2$ control parameters of a Beta distribution to allow sampling actions within a truncated range \citep{Chou2017}.
We shift and scale the sampled values to correspond to the task action range.
We also include entropy regularization to prevent the policy from becoming overly deterministic early during training.

\begin{table}[h!]
    \caption{Implementation details for experiments using PPO}
    \label{tab:ppo-details}
    \begin{center}
    \renewcommand{\arraystretch}{1.4}
    \begin{tabular}{r|c|c|c|c|c|c|c|c|c}
    & \multicolumn{3}{|c}{Point maze} & \multicolumn{3}{|c}{Ant maze (high)} & \multicolumn{3}{|c}{Bit flipping}\\
    Hyperparameter & PPO & +\mthd & +ICM & PPO & +\mthd & +ICM & PPO & +\mthd & +ICM\\
    \toprule
    Rollouts per Update & \multicolumn{3}{|c|}{4} & \multicolumn{3}{c}{4} & \multicolumn{3}{|c}{4}\\
    \hline
    Epochs per Update & \multicolumn{3}{|c|}{4} & \multicolumn{3}{c}{2} & \multicolumn{3}{|c}{4}\\
    \hline
    m.Batches per Epoch & \multicolumn{3}{|c|}{4} & \multicolumn{3}{c}{4} & \multicolumn{3}{|c}{4}\\
    \hline
    Learning Rate (LR) & \multicolumn{3}{|c|}{0.001}  & \multicolumn{3}{c}{0.001}  & \multicolumn{3}{|c}{0.001}\\
    \hline
    LR Decay & \multicolumn{3}{|c|}{0.999}  & \multicolumn{3}{c}{1.0} & \multicolumn{3}{|c}{0.999}\\
    \hline
    Entropy Reg $\lambda$ & \multicolumn{3}{|c|}{0.025} & \multicolumn{3}{c|}{0.025} & 0.025 & 0.0 & 0.025\\
    \hline
    GAE $\lambda$  & \multicolumn{3}{|c|}{0.98}  & \multicolumn{3}{c}{0.98}  & \multicolumn{3}{|c}{0.98}\\
    \hline
    Bootstrap Value & \multicolumn{2}{|c|}{N} & Y & \multicolumn{2}{|c|}{N} & Y & \multicolumn{2}{|c|}{N} & Y\\
    \hline
    Discount Factor & \multicolumn{2}{|c|}{1.0} & 0.98 & \multicolumn{2}{|c|}{1.0} & 0.98 & \multicolumn{2}{|c|}{1.0} & 0.98\\
    \hline
    Inclusion thresh. ($\hp$) & & 5.0 & & & 10.0 & & & Inf &\\
    \end{tabular}
    \end{center}
\end{table}

\begin{table}[h!]
    \caption{Implementation details for off-policy experiments}
    \label{tab:ppo-details}
    \begin{center}
    \renewcommand{\arraystretch}{1.4}
    \begin{tabular}{r|c|c|c}
    Hyperparameter & Point maze & Ant maze (high) & Bit flipping\\
    \toprule
    Rollouts per Update & \multicolumn{3}{|c}{4}\\
    \hline
    m.Batches per Update & \multicolumn{3}{|c}{40}\\
    \hline
    m.Batches size  & 64 & 128 & 128\\
    \hline
    Learning Rate (LR) & \multicolumn{3}{|c}{0.001}\\
    \hline
    Action $L_2 ~ \lambda$ & 0.25 & 0.0002 & NA\\
    \hline
    Behavior action noise  & \multicolumn{2}{|c|}{$0.1 ~\times $ ~action range} & NA\\
    \hline
    Behavior action epsilon  & \multicolumn{3}{|c}{0.2}\\
    \hline
    Polyak coefficient  & \multicolumn{3}{|c}{0.95}\\
    \hline
    Bootstrap Value & \multicolumn{3}{|c}{Y}\\
    \hline
    Discount Factor & \multicolumn{3}{|c}{0.98}\\
    \end{tabular}
    \end{center}
\end{table}

\paragraph{Intrinsic Curiosity Module (ICM).}
We base our implementation of ICM off the guidelines provided in \citet{Burda2018a}.
We weigh the curiosity-driven intrinsic reward by 0.01 compared to the sparse reward.
Note that in the settings we used, ICM is only accompanied by sparse extrinsic rewards, meaning that it only experiences the intrinsic rewards until it (possibly) discovers the goal region.
During optimization, we train the curiosity network modules (whose architectures follow similar designs to the policy and value for the given task) at a rate of 0.05 compared to the policy and value network modules.


\begin{table}[h!]
    \caption{Environment details}
    \label{tab:env-details}
    \begin{center}
    \renewcommand{\arraystretch}{1.4}
    \begin{tabular}{c|c|c|c}
    \textbf{Setting} & $\State \in$ & $\Goal \in$ & $\Action \in$\\
    \toprule
    Point maze & $\Rscalar^2$ & $\Rscalar^2$ & $[-0.95, 0.95]^2$\\
    \hline
    Ant maze (high) & $\Rscalar^{30}$ & $\Rscalar^2$ & $[-5, 5]^2$\\
    \hline
    Ant maze (low) & $\Rscalar^{30}$ & $\Rscalar^2$ & $[-30, 30]^8$\\
    \hline
    Bit flipping & $\{0, 1\}^{13\times13\times2}$ & $\{0, 1\}^{13\times13}$ & $\{0...9\}$\\
    \hline
    \multirow{2}{*}{Minecraft} & $\state^v \in \Rscalar^{80\times120\times3},$ & \multirow{2}{*}{$\{0...N_{b}\}^{11\times11\times3}$} & \multirow{2}{*}{$\{0...20\}$}\\
    & $\state^c \in \{0...N_{b}\}^{11\times11\times3}$ & &\\
    \end{tabular}
    \end{center}
\end{table}

\begin{table}[h!]
    \caption{Task details}
    \label{tab:task-details}
    \begin{center}
    \renewcommand{\arraystretch}{1.2}
    \begin{tabular}{c|c|c|c|c}
    \textbf{Setting} & $\sgmap{\state}$ & $\distance{}{}$ & $\delta$ & Max. $T$ \\
    \toprule
    Point maze & $\mathbb{I}$ & $L_2$ & $0.15$ & 50\\
    \hline
    Ant maze (high) & $[\state^0, \state^1]$ & $L_2$ & $1.0$ & 25 (=500 env steps)\\
    \hline
    Ant maze (low) & $[\state^0, \state^1]$ & $L_2$ & NA & 20 (env steps)\\
    \hline
    Bit flipping & $\state^{\colon,\colon,0}$ & $L_1$ & $0.0$ & 50\\
    \hline
    Minecraft & $\state^c$ & $\sum x_{ijk} \neq y_{ijk}$ & $0.0$ & 100\\
    \end{tabular}
    \end{center}
\end{table}

\paragraph{2D point maze navigation.}
The 2D point maze is implemented in a 10x10 environment (arbitrary units) consisting of an array of pseudo-randomly connected 1x1 squares.
The construction of the maze ensures that all squares are connected to one another by exactly one path.
This is a continuous environment.
The agent sees as input its 2D coordinates and well as the 2D goal coordinates, which are always somewhere near the top right corner of the maze.
The agent takes an action in a 2D space that controls the direction and magnitude of the step it takes, with the outcome of that step potentially affected by collisions with walls.
The agent does not observe the walls directly, creating a difficult exploration environment.
For all experiments, we learn actor and critic networks with 3 hidden layers of size 128 and \texttt{ReLU} activation functions.

\paragraph{Ant maze navigation with hierarchical RL.}
The ant maze experiment borrows a similar set up to the point maze but trades complexity of the maze for complexity in the navigation behavior.
We use this as a lens to study how the different algorithms handle HRL in this setting.
We divide the agent into a high-level and low-level policy, wherein the high-level policy proposes subgoals and the low-level agent is rewarded for reaching those subgoals.
For all experiments, we allow the high-level policy to propose a new subgoal $\goal^L$ every 20 environment timesteps.
From the perspective of training the low-level policy, we treat each such 20 steps with a particular subgoal as its own mini-episode.
At the end of the full episode, we perform 2 epochs of PPO training to improve the low-level policy, using distance-to-subgoal as the reward.

The limits of the maze are $[-4, 20]$ in both height and width.
The agent starts at position $(0, 0)$ and must navigate to goal location $\goal=(x_\goal, y_\goal)$ with coordinates sampled within the range of $x_\goal\in[-3.5, 3.5]$ and $y_\goal\in[12.5, 19.5]$.
It should be noted that, compared to previous implementations of this environment and task \citep{Nachum2018}, we do not include the full range of the maze in the distribution of task goals.
For the agent to ever see the sparse reward, it must navigate from one end of the U-maze to the other and cannot bootstrap this exploration by learning from goals that occur along the way.
As one might expect, the learning problem becomes considerably easier when this broad goal distribution is used; we experiment in the more difficult setting since we do not wish to impose the assumption that a task's goal distribution will naturally tile goals from ones that are trivially easy to reach to those that are difficult.

At timestep $t$, the high-level controller outputs a 2-dimensional action $\action_t \in [-5, 5]^2$, which is used to compute the subgoal $g^L_t = \sgmap{\state_t} + \action_t$.
In other words, the high-level action specifies the relative coordinates the low-level policy should achieve.
From the perspective of training the high-level policy, we only consider the timesteps where it takes an action and consider the result produced by the low-level policy as the effect of having taken the high-level action.

In all experiments, both the high- and low-level actor and critic networks use 3 hidden layers of size 128 and \texttt{ReLU} activation functions.

\paragraph{2D bit flipping task.}
We extend the bit flipping example used to motivate HER \citep{Andrychowicz2017} to a 2D environment in which interaction with the bit array depends on location.
In this setting, the agent begins at a random position on a 13x13 grid with none of its bit array switched on.
Its goal is to reproduce the bit array specified by the goal.
To populate these examples, we procedurally generate goal arrays by simulating a simple agent that changes direction every few steps and toggles bits it encounters along the way.

We include this example mostly to illustrate (i) that our method can work in this entirely discrete learning setting and (ii) that naive distance-to-goal based rewards are exceptionally prone to even brittle local optima, such as the ones created when the agent learns to avoid taking the toggle-bit action.

We report the (eventually) successful performance using vanilla DQN but point out that this required modifying the reward delivery for this particular agent.
In all previous settings, agents trained on shaped rewards receive that reward only at the end of the episode (and no discounting is used).
While it is beyond the scope of this work to decipher this observation, we found that DQN could only learn if the shaped reward was exposed at every time step (using a discounting of $\gamma=0.98$).
The variant that used the reward-at-end scheme never learned.

For all bit flipping experiments, we use 2D convolution to encode the states and goals.
We pool the convolution output with \texttt{MaxPooling}, apply \texttt{LayerNorm}, and finally pass the hidden state through a fully connected layer to get the actor and critic outputs.

\paragraph{3D construction in Minecraft.}
To test our proposed method at a more demanding scale, we implement a custom structure-building task in Minecraft using the Malmo platform.
In this task, we place the agent at the center of a ``build arena'' which is populated in one of several full Minecraft worlds.
In this particular setting, the agent has no task-specific incentive to explore the outer world but is free to do so.
Our task requires the agent to navigate the arena and control its view and orientation in order to reproduce the structure provided as a goal (similar to a 3D version of the bit flipping example but with richer mechanics and more than one type of block that can be placed).
All goals specify a square structure made of a single block type that is either 1 or 2 blocks high with corners at randomly chosen locations in the arena.
For each sampled goal, we randomly choose those configuration details and keep the sampled goal provided that it has no more than 34 total blocks (to ensure that the structure can be completed within a 100 timestep episode).
The agent begins each episode with the necessary inventory to accomplish the goal.
Specifically, the goal structures are always composed of 1 of 3 block types and the agent always begins with 64 blocks of each of those types.
It may place other block types if it finds them.

The agent is able to observe the first-person visual input of the character it controls as well as the 3D cuboid of the goal structure and the 3D cuboid of the current build arena.
The agent therefore has access to the structure it has accomplished but must also use the visual input to determine the next actions to direct further progress.

The visual input is process through a shallow convolution network.
Similarly, the cuboids, which are represented as 3D tensors of block-type indices, are embedded through a learned lookup and processed via 3D convolution.
The combined hidden states are used as inputs to the policy network.
The value network uses separate weights for 3D convolution (since it also takes the anti-goal cuboid as input) but shares the visual encoder with the policy.

Owing to the computational intensity and long run-time of these experiments, we limit our scope to the demonstration of \method ~in this setting.
However, we do confirm that, like with the bit flipping example, naive distance-to-goal reward shaping fails almost immediately (the agent learns to never place blocks in the arena within roughly 1000 episodes).

For the work presented here, we compute the reward as the change in the distance produced by placing a single block (and use discounting of $\gamma=0.99$).
We find that this additional densification of the reward signal produces faster training in this complex environment.

\end{document}